# Integrating Three Mechanisms of Visual Attention for Active Visual Search

Amir Rasouli, John K. Tsotsos
Dept. of Electrical Engineering and Computer Science and Center for Vision Research
York University, Toronto, Canada
{aras, tsotsos}@cse.yorku.ca

*Abstract*—Algorithms for robotic visual search can benefit from the use of visual attention methods in order to reduce computational costs. Here, we describe how three distinct mechanisms of visual attention can be integrated and productively used to improve search performance. The first is viewpoint selection as has been proposed earlier using a greedy search over a probabilistic occupancy grid representation. The second is top-down object-based attention using a histogram backprojection method, also previously described. The third is visual saliency. This is novel in the sense that it is not used as a region-of-interest method for the current image but rather as a non-combinatorial form of look-ahead in search for future viewpoint selection. Additionally, the integration of these three attentional schemes within a single framework is unique and not previously studied. We examine our proposed method in scenarios where little or no information regarding the environment is available. Through extensive experiments on a mobile robot, we show that our method improves visual search performance by reducing the time and number of actions required.

## I. Introduction

The ability to search an environment, whether to look for a particular object or to explore an unknown area is a fundamental requirement for any autonomous robot. In either case the robot has to select viewpoints from which to observe the environment. Here the efficiency of viewpoint selection is vital to the system because the resources such as battery energy or time are often limited.

A trivial approach to visual search is simply to look from every possible viewpoint within an environment. However, it is proven that without the use of attentive processes the problem of search is NP-hard. This means that it has exponential time complexity independent of the implementation [1]. Hence, brute force search is not feasible in any practical system.

The psychophysical literature presents numerous arguments in favor of visual attention and its necessity in animal and machine [2, 3]. Visual saliency is a commonly used attentional mechanism in robotics. Butko *et al.* [4] use saliency to identify humans based on detected motion patterns to enable a social robot to detect people and interact with them. In [5] a humanoid robot builds saliency maps by tracking an instructor's gaze while he is performing various tasks and later uses these maps to imitate instructor's actions. Orabona *et al.* [6] apply saliency to object recognition. A saliency map is computed as a linear combination of individual color channel responses and the area in the map with the highest saliency is assumed to be the target. In [7] the authors generate a bottom-up saliency map using a series of low level features such as color, intensity and orientation. This map is used to select features which help the robot to localize itself within an environment. In a similar context, Kim and Eustice [8] employ saliency to select key-frames for localizing an underwater autonomous robot. The saliency map is generated using an online bag-of-words approach and is used to discard featureless and unsuitable images. Robert *et al.* [9] propose a motion based saliency model for the fast detection of trees and use it to navigate an aerial robot in forests.

In the context of visual search, attention is used for efficient gaze control and viewpoint selection. Garvey [10] proposes the notion of indirect search. He divides the task of search into two stages of identifying an intermediate object that is spatially related to the object of interest and restricting the search to those regions. Aydemir *et al.* [11] and later Gobekbecker *et al.* [12] apply this notion in practice by computing a probability distribution of target presence relative to intermediate objects based on predefined spatial relations between them. The major drawback of these approaches is that searching for an intermediate object is not necessarily simpler than finding the target directly. Ye and Tsotsos use attention in visual search in the form of Bayesian reasoning [13]. In their model every location in the search environment is assigned a probability of the target being present. Then the task of search is to choose the direction that yields the highest probability of detecting the target subject within a maximum time constraint. They propose a greedy algorithm to select the next best action. Andreopoulos *et al.* [14] use a similar Bayesian framework to search for an object in a 3D environment. At each point of the search, from a list of all possible locations and gazes the robot greedily selects the best combination of location and gaze that maximizes the chance of detecting the target in the shortest time possible.

The contribution of this paper is twofold. First we propose a three-pronged probabilistic search algorithm and through experimental evaluations on a mobile robot show what improvements can be achieved by incorporating three forms of visual attention into visual search. Second we introduce an extension to the method in [13] by removing the predefined environment boundaries and demonstrate the effect of this change on the overall performance.

## II. ATTENTION FOR VIEWPOINT SELECTION

Our viewpoint selection mechanism is adopted from [13]. In this search model, the robot is looking for an object in an unknown 3D environment with known exterior boundaries. The search region $\Omega$ is tessellated into a 3D grid of non-overlapping cubic elements (an occupancy grid), $c_i$, $i = 1 \ldots n$ each holding the probability value of the target presence. Here the search agent's action is defined by an operation $f = f(S(\tau), a)$, on $\Omega$, which consists of taking an image with known camera configuration $S(\tau)$ and processing it by algorithm $a$ to detect the target, where $S(\tau)$ specifies the camera position $(x_c, y_c, z_c)$, direction of viewing axis $(p, t)$, and the width and height of its solid viewing angle $(w, h)$ at time $\tau$. Performing each action $f$ incurs a cost $t(f)$, which accounts for every aspect of operation such as acquiring and processing an image.

Let $p((x, y, z), \tau)$ denote the probability of the target at location $(x, y, z)$ at time $\tau$ and $p(c_{out}, \tau)$ be the probability of the target to be outside the search region $\Omega$ at time $\tau$. The probability of detecting the target by applying operation $f$ is given by,

$$P_{\Psi_f}(f) = \sum_{c_i \in \Psi_f} p(c_i, \tau_f) b(c_i, f), \quad (1)$$

where $\tau_f$ denotes the time just before $f$ is applied and $\Psi_f$ is the influence range of the action $f$, i.e. regions within $\Omega$ that are visible to the search agent with the current camera setting $S(\tau)$. Here $b(c_i, f)$ is the detection function that specifies the conditional probability of detecting the target by applying action $f$ given that the target is centered at cube $c_i$.

Let $O_\Omega$ denote the set of all possible operations on $\Omega$. Then the effort allocation $F = \{f_1, \ldots, f_k\}$, $f_i \in O_\Omega$, is an ordered set of operations applied to perform the visual search. We define the task of search as the effort allocation $F$ that maximizes the probability of detecting the target subject to $\sum_{f \in F} t(f) < K$ where $K$ is the total cost allowable for the search. In [13] the viewpoint selection is controlled by probability $P_{\Psi_f}(f)$, i.e. the robot attends the locations first that have the highest probability of detecting the target. Ye and Tsotsos [13] later show that for a practical system a greedy algorithm suffices as a good approximation to the problem of viewpoint selection.

They divide the search process to two stages of "where to look next" and "where to go next". At the "where to look next" stage, a 'best-first' strategy is employed to examine all possible actions at the current location. The goal is to select an operation $f$ that yields the highest utility given by $E_{\Psi_f}(f) = \frac{\sum_{c_i \in \Psi_f} p(c_i, \tau_f) b(c_i, f)}{t(f)}$. Once an operation is performed, the target's location probabilities are updated as follows:

$$p(c_i, \tau_{f+}) = \frac{p(c_i, \tau_f)(1 - b(c_i, f))}{p(c_{out}, \tau_f) + \sum_{j=1}^{n} p(c_j, \tau_f)(1 - b(c_j, f))},$$

$i = 1, \ldots, n, out$, (2)

where $\tau_{f+}$ is the time after $f$ is applied and $p(c_{out}, \tau_{f+})$ is the probability that the target is outside the search region $\Omega$ at time $\tau_{f+}$. Intuitively, if the robot cannot find the target by performing operation $f$, the probability of the influence range (the range within which the recognition algorithm can detect the target) decreases as the other regions' probabilities increase.

Once the covering probability of all remaining operations $Prob_{\Psi_f} = \sum_{c_i \in \Psi_f} p(c_i)$ goes below some threshold $\Theta_{move}$, the robot moves to a different location. During "where to move next" stage the robot selects the location with the highest probability of detecting the target given by $Prob_{\Psi_j} = \sum_{c_i \in \Psi_j} p(c_i)$, where $\Psi_j$ is the region within the union of all influence ranges of view at position $j$.

In the original formulation of search in [13] attention is only used as a viewpoint selection mechanism. At each stage of search the robot relies only on a recognition algorithm to update the likelihood of finding an object. However, in practice the spatial range of recognition algorithms is typically lower than the range of disparity detection by stereo cameras. So if we can find clues regarding the target's presence in regions beyond the effective range of the recognition algorithm, we can locate them within the environment and search those regions first. For this purpose in the following sections we introduce two attentive mechanisms namely visual saliency and object-based attention.

## III. VISUAL SALIENCY IN VISUAL SEARCH

In the computer vision literature visual saliency is commonly used to identify conspicuous parts of an image. The most common form of visual saliency is bottom-up, which is entirely data-driven without any object-specific knowledge [15]. Although bottom-up models do not use specific knowledge, they still can be effective in the context of search. The saliency results generated by these models can potentially be used to find objects that stand out in an environment or physical structures spatially related to the target. Bottom-up saliency maps are typically computed as a combination of low level feature maps such as color and intensity [7], superpixels [16], shapes [17], etc. Alternatively, there are biologically inspired approaches to generating features such as sparse coding [18] and ICA [19].

For bottom-up saliency we chose Attention based on Information Maximization (AIM) [20] algorithm for its superior performance in different visual contexts (see [21] for more details). The AIM algorithm first convolves an image with a set of filters computed by applying the Independent Component Analysis (ICA) [19] to a large number of indoor and natural image samples. Then, the joint likelihood of filter responses is calculated using a Gaussian window as follows:

$$p(w_{i,j,k} = v_{i,j,k}) = \frac{1}{\sigma\sqrt{2\pi}} \sum_{\forall s,t \in \Psi} \omega(s,t) e^{-(v_{i,j,k} - v_{i,s,t})^2 / 2\sigma^2}, \quad (3)$$

with $\sum_{s,t} \omega(s,t) = 1$, where $w_{i,j,k}$ denotes a set of independent coefficients based on a neighborhood centered at $j$ and $k$, $v_{i,j,k}$ represents local statistics and $\Psi$ is the context on which the probability estimate of the coefficients of $\omega$ is based. Given the statistical independence of ICA generated features, the overall probability density function of features can be computed by

$$p(w_1 = v_1, w_2 = v_2, \ldots, w_n = v_n) = \prod_{i=1}^{n} p(w_i = v_i). \quad (4)$$

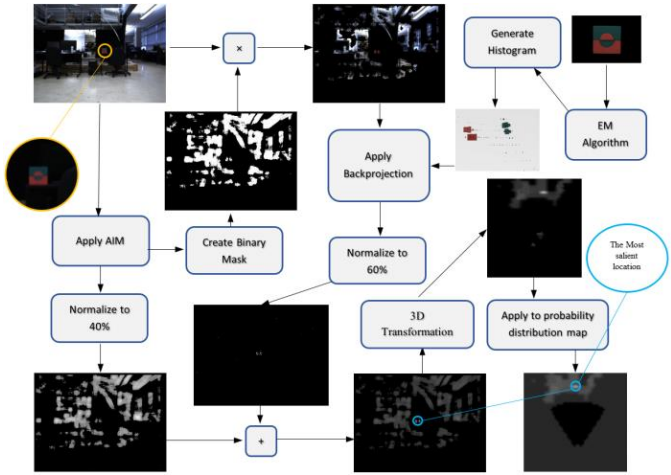

Figure 1. The process of applying saliency to the robotic visual search.

To measure the saliency of each local neighborhood, the self-information measure of each distribution is computed as a sum of negative log-probabilities of each filter response. The regions with higher values of self-information measure are statistically rare within the image and are recognized as salient.

IV. OBJECT-BASED ATTENTION IN OBJECT SEARCH

As the name implies, object-based attentive mechanisms are used to identify image regions related to an object. These models commonly use data-driven approaches (e.g. bottom-up saliency) as a preprocessing step and then use object specific knowledge to find salient regions within an image. Object-based attention techniques generally rely on some form of training to learn object-specific features, e.g. SVM [22] and EM [23]. In our work we use histogram backprojection (HB) [24]. This method does not require offline training and is computationally efficient.

The first step in HB is to generate a 3D histogram of an object's RGB color distributions. For this purpose a target template is generated using an Expectation Maximization (EM) model [25] to separate the foreground (the target) from the background. The template is pixel-wise normalized to reduce the effects of illumination changes.

Let $h(C)$ be the histogram function which maps color $C = (red, green, blue)$ to a bin of histogram $H(C)$ computed from the object's template. The backprojection of the object over the image is given by,

$$\forall x, y: b_{x,y} := h(I'_{x,y,c}), \quad (5)$$

where $b$ is the grayscale backprojection image and $I'$ is the normalized image $I$.

A linear weighted combination of AIM saliency and HB outputs forms the final conspicuity map. After each unsuccessful attempt to detect the target, the last captured image is further processed by applying AIM. The resulting saliency map is then thresholded at a predefined percentile value (in our work 80%). A binary version of the AIM map is also applied to the original image in the form of a mask to extract RGB values of the interest regions,

$$\hat{I}_\theta = I_\theta \times M(x, y),$$

$$\begin{cases} M(x, y) = 1 & info(x, y) > p \\ M(x, y) = 0 & else, \end{cases} \quad (6)$$

where $I_\theta$ is the original image captured with camera configuration $\theta$, $info(x, y)$ is the result of AIM, $M(x, y)$ is the binary mask and $p$ denotes the percentile threshold. The image $\hat{I}_\theta$ is used to generate backprojection map, using a predefined 3D color histogram of the target's template. The final saliency map is a combination of AIM (40%) and HB (60%).

With the aid of a stereo camera, the 3D coordinates of the salient locations are calculated and mapped to the 3D grid of the search environment. If salient regions fall within the effective range of the recognition algorithm, they are ignored, otherwise saliency values are used to update the probability distribution of target's location. Figure 1 illustrates the entire process.

V. ATTENTIVE VISUAL SEARCH IN ENVIRONMENTS WITH UNKNOWN BOUNDARIES

In the original formulation of search in [13] the probability values are assigned to the environment based on the assumption that the external boundaries of the environment are known. However, in practice this is not always the case. For instance, in exploration and mapping applications the environment is completely unknown and boundaries are detected as the robot proceeds. This raises the question of viewpoint selection in such scenarios.

The initial search region is the size of the viewing sphere of the robot with radius dependent on the range of accurate disparity estimation of the camera system and the performance radius of the recognition algorithm. The robot starts by searching its surroundings within its viewing sphere. It establishes a primary 3D map of the environment and assigns the probabilities of the cubic elements according to (2). If the target was not found from the initial position, the robot follows the same "look first, move next" strategy. With each change of position, the search boundaries are adjusted accordingly, i.e. the total 3D region is determined by the union of the new and previous viewing spheres. Note that the search region does not extend beyond physical obstacles, which are not accessible by the robot, such as walls.

VI. EXPERIMENTS

We implemented the described methods of search on a Pioneer 3, a four-wheeled differential-drive mobile robot. The robot is equipped with a Point Gray Bumblebee stereo camera mounted on a Directed Perception pan-tilt unit. Three office environments of various sizes and with different furniture configurations were used. The environments were tessellated into voxels of size $50^3 mm^3$. We set threshold $\Theta_{move}$ empirically to 0.05. The recognition algorithm in our model was based on normalized gray-scale correlation algorithm described in [26]. This algorithm is not view-independent and the target of interest only is recognized when facing toward the camera up to 45° degree of transformation in depth axis.

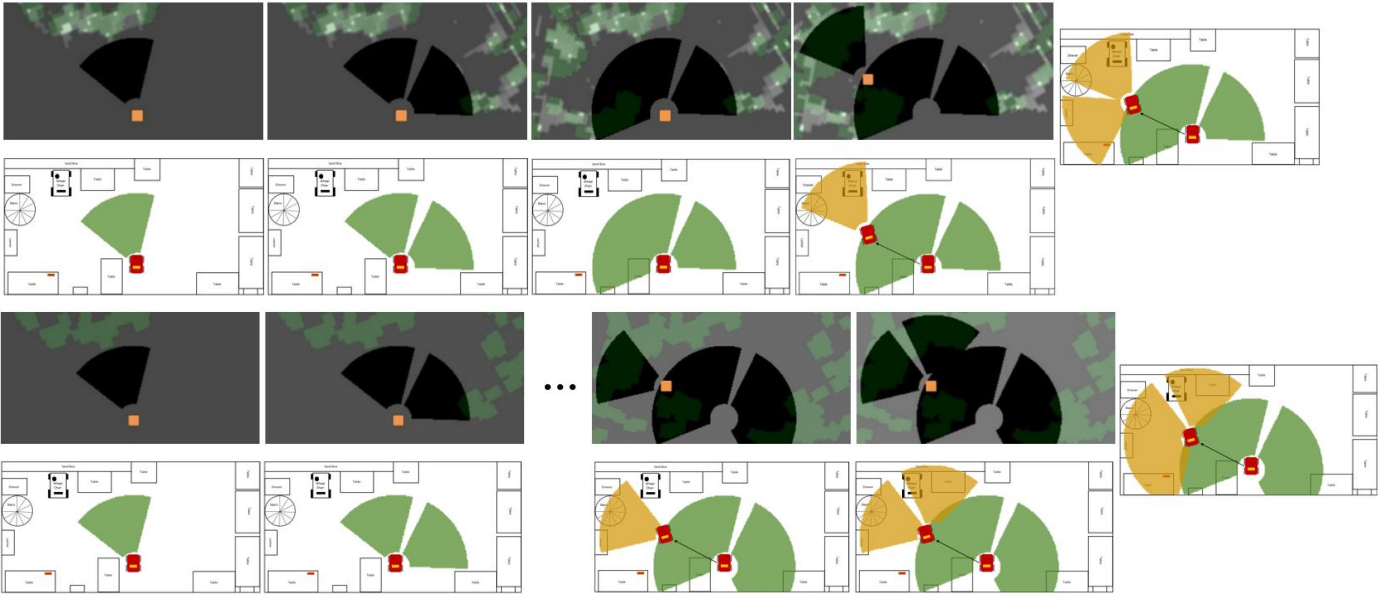

*Figure 2. The search without (on the top) and with (at the bottom) the use of saliency. The grayscale images correspond to the 2D representation of the probability distribution maps. Here the green regions are obstacles, the yellow square is the robot and white areas (in the saliency example) are the salient regions observed by the robot. In the top view images, the small red rectangle at the bottom is the target and the larger one is the robot.*

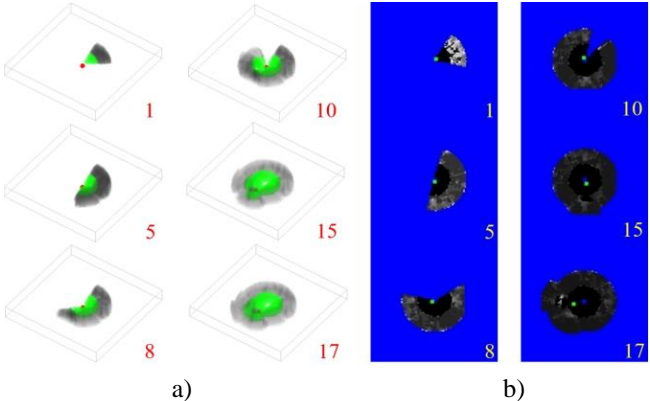

*Figure 3. An example of 3D search without boundaries. a) the 3D cloud of probability map. The red square is the robot, the green regions are the areas with probability value zero and gray points represent the probability of the target presence. b) the 2D representation of the probability map. Blue color represents unknown regions, the green square is the robot and gray points show the probability values of the environment.*

### A. Three-Pronged Attentive Visual Search

The first set of experiments aims to evaluate the benefit of combining the above attention mechanisms in visual search. We conducted a total of 106 experiments by placing the robot and the target in random locations. We ran the search algorithm with and without the use of saliency and object-based models (see Figure 2 for an example). Table 1 summarizes the results of our experiments. These results are divided into two groups of No Move (NM) in which the object was found before the robot moves to a new location, and Move (M) in which the robot moved at least once to find the object.

TABLE I. THE AVERAGE RESULTS OF EXPERIMENTS CONDUCTED IN 3 ENVIRONMENTS

| Method | Factor | Search Process | | |
|---|---|---|---|---|
| | | *NM* | *M* | *Overall* |
| **Search with no Attention** | *No. Actions* | 2.03 | 10.50 | 8.30 |
| | *Time(min)* | 2.19 | 19.33 | 14.97 |
| | *Distance Travelled (m)* | 0 | 11.85 | 8.91 |
| **Search with Attention** | *No. Actions* | 2.03 | 8.49 | 6.79 |
| | *Time (min)* | 2.23 | 15.03 | 11.68 |
| | *Distance Travelled (m)* | 0 | 9.81 | 7.28 |

Saliency and object-based information corresponds to regions beyond the range of the recognition algorithm, thus, the robot has to move in order to observe the salient regions. As a result the performance of the robot is improved significantly in Move scenarios. In fact, our proposed method found the object faster in 77% of the cases than the method in [13] in terms of the number of actions that are required to find the object. As expected, in the cases of No Move both methods of search performed identically.

Note that saliency also can be distracting in the context of search. In some cases it diverted the attention of the robot away from the target. This particularly was true in situations when the target was placed in areas with more salient objects.

### B. Attentive Search without Boundary Information

We conducted a number of experiments to compare the performance of our attention model in scenarios with and without boundary information. The experiments were done in the same environments and setup as before. A 3D example of the search with unknown boundaries is shown in Figure 3. Table 2 shows the results of these experiments.

TABLE II. THE AVERAGE RESULTS UNKNOWN SEARCH EXPERIMENTS

| Method | Known Boundaries | Unknown Boundaries |
|---|---|---|
| *No. Actions* | 8.5 | 19 |
| *Time(s)* | 417.44 | 892.78 |
| *Distance Travelled (m)* | 3.07 | 3.77 |

As it is expected, our attentive search model is less effective when the boundaries are unknown. Here the performance is reduced to less than half comparing to scenarios with known boundary information. This is due to following factors: first generating the primary map is time consuming and it typically involves performing up to 12 viewpoint selections and accompanying processing. Second, since the borders of the environment are unknown, the robot moves more conservatively, i.e. it takes smaller steps to get to a new location, which increases the overall traveling distance to expand the region.

## VII. CONCLUSION

In this paper we presented a novel attentive search algorithm in which three visual attention mechanisms, namely viewpoint selection, saliency and object-based models are combined. In our model attention is used to generate maps highlighting areas in the image which are more likely to contain an object of interest. With the aid of a viewpoint selection approach the attention of the robot is directed to those interest points.

Through experimental evaluations we showed that by using the proposed three-tier attention framework we can significantly improve visual search by decreasing the cost in terms of the number of actions taken, the search time and the distance travelled by the robot.

We also demonstrated how knowledge about the search environment can influence the effectiveness of attention in search. Our experimental results show that removal of the boundary information reduces the efficiency of the search algorithm by more than a half.

The proposed attention model was evaluated in small office environments, where dimensions of each location did not significantly exceed the effective range of the robot's recognition algorithm. We anticipate that incorporating attention in visual search will have a more prominent effect in larger environments.

The object-based model we used in our work currently relies only on the color distribution features. To reduce the number of false detections it can be extended by adding additional object features such as shape and orientation.

## ACKNOWLEDGMENT

We acknowledge the financial support of the Natural Sciences and Engineering Research Council of Canada (NSERC), the NSERC Strategic Network for Field Robotics, and the Canada Research Chairs Program through grants to JKT. We also thank Yulia Kotseruba for her assistance and helpful feedback on the project.